\pdfoutput=1

\documentclass[a4paper,11pt]{article}

\usepackage[preprint]{acl}

\usepackage{times}
\usepackage{latexsym}

\usepackage[T1]{fontenc}
\usepackage[utf8]{inputenc}
\usepackage{multirow}

\usepackage{microtype} 

\usepackage{inconsolata} 

\usepackage{subfiles}

\usepackage{booktabs} 
\usepackage{enumitem}

\usepackage{amsmath}
\usepackage{amsthm}
\usepackage{amssymb}

\usepackage[capitalise]{cleveref}

\usepackage{graphicx}

\usepackage{xcolor}
\usepackage{hyperref}

\usepackage{caption}

\usepackage{siunitx} 
\usepackage{comment}

\usepackage{csquotes} 

\theoremstyle{definition}

\newcommand{\ee}{\mathrm{e}}   
\renewcommand{\Pr}{\mathbb{P}} 
\newcommand{\EE}{\mathbb{E}}   
\newcommand{\VV}{\mathbb{V}}   

\definecolor{myblue}{RGB}{20,80,150}  
\definecolor{myred}{RGB}{160,30,30}   
\definecolor{mygreen}{RGB}{50,120,50} 

\usepackage{pifont}
\newcommand{\affone}{\color[HTML]{153268}\ding{117}}
\newcommand{\afftwo}{\color[HTML]{000000}\ding{60}}
\newcommand{\affthree}{\color[HTML]{006c66}\ding{118}} 

\title{Vocabulary Shapes Cross-Lingual Variation of \\ Word-Order Learnability in Language Models}

\author{
    \textbf{Jonas Mayer Martins}\textsuperscript{\affone} ~~~~
    \textbf{Jaap Jumelet}\textsuperscript{\afftwo} ~~~~
    \textbf{Viola Priesemann}\textsuperscript{\affone,\affthree} ~~~~
    \textbf{Lisa Beinborn}\textsuperscript{\affone} \\
    \textsuperscript{\affone} University of G\"{o}ttingen, Germany ~~~~
    \textsuperscript{\afftwo} University of Groningen, Netherlands \\ 
    \textsuperscript{\affthree} MPI for Dynamics and Self-Organization, Germany \\ 
    \texttt{firstname.lastname@uni-goettingen.de} \\}

\begin{document}

\maketitle

\begin{abstract}
Why do some languages like Czech permit free word order, while others like English do not? We address this question by pretraining transformer language models on a spectrum of synthetic word-order variants of natural languages. We observe that greater word-order irregularity consistently raises model surprisal, indicating reduced learnability. Sentence reversal, however, affects learnability only weakly. A coarse distinction of free- (e.g., Czech and Finnish) and fixed-word-order languages (e.g., English and French) does not explain cross-lingual variation. Instead, the structure of the word and subword vocabulary strongly predicts the model surprisal. Overall, vocabulary structure emerges as a key driver of computational word-order learnability across languages.
\end{abstract}

\begin{center}
\small
{
\href{https://gitlab.gwdg.de/huds/projects/shuffle/-/tree/v1.0.2}
    {
        \raisebox{-0.33em}{\includegraphics[height=1.2em]{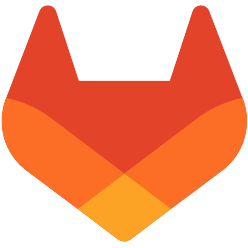}} 
        \;Code repository
    }
}
\end{center}


\section{Introduction} 

Human languages have emerged over millennia through dynamics shaped by communicative and cognitive constraints \cite{zipf-1935-psychobiology,piantadosi-etal-2012-communicative,hawkins-2014-crosslinguistic,futrell-etal-2020-lossycontext,hahn-xu-2022-crosslinguistic,clark-etal-2023-crosslinguistic}. Yet, within those universally shared bounds, languages exhibit a striking typological diversity, varying in morphological complexity and preferred word orders, for example. Languages, in all their diversity, are not equally complex in every aspect \cite{croft-2002-typology,sampson-etal-2009-language,koplenig-etal-2023-large}. This raises a central question: Are all languages equally hard to learn? And if not, why?

One dimension of linguistic diversity is word-order flexibility---the degree to which words in a sentence can be reordered without changing its meaning, except for emphasis.
In Czech, for instance, case marking determines the grammatical role of nouns in a sentence, allowing constituent order to vary relatively freely. In the sentence \enquote{Robot maluje ko\v{c}ku.} (\emph{The robot paints the cat.}), any of the six permutations of subject (robot), verb (maluje), and object (ko\v{c}ku, the accusative case of ko\v{c}ka) is grammatically acceptable and conveys the same core meaning. In English, by contrast, the sentence \enquote{The robot paints the cat.} cannot be reordered without changing its meaning or rendering it ungrammatical.

\paragraph{Research questions}
The Czech--English example illustrates a general typological pattern: Languages with relatively free word order (like Czech) tend to encode syntactic relations through morphology, while languages with relatively fixed word order (like English) rely on word position instead. This contrast motivates two questions: First, whether learnability is sensitive to the degree of word-order flexibility; and second, why some languages are more robust to free word order than others. Instead of the historical emergence of word order, these research questions target learnability as an inherent property of languages.

\paragraph{Synthetic languages}
In natural languages, typological features are often strongly correlated \cite{greenberg-1990-universals}. Synthetic-language experiments aim to solve this problem by perturbing a natural language along a single dimension, for example altering word order while preserving vocabulary and content \cite{kallini-etal-2024-mission,xu-etal-2025-can,yang-etal-2025-anything}. However, prior work has faced two limitations: First, word order flexibility and morphological complexity are mostly studied in isolation, although these two factors are clearly connected \cite{bisazza-etal-2021-difficulty,nijs-etal-2025-word,liu-etal-2025-complexity}. Perturbation experiments commonly operate at the subword level, which often breaks up lexical units. For example, a subword tokenizer might split the Czech word \emph{maluje} into \emph{ma} and \emph{luje}, which yields linguistically implausible sequences when shuffled. Thus, word order and morphology are perturbed simultaneously. Second, the use of disparate shuffling methods with discrete parameters limits control over the perturbation strength and makes it difficult to compare results. Due to these limitations, the interplay of word-order flexibility, morphological complexity, and tokenization in shaping computational learnability remains an open question \cite{arnett-bergen-2025-why,poelman-etal-2025-confounding}.

\begin{figure}
    \centering
    \includegraphics[width=1\linewidth]{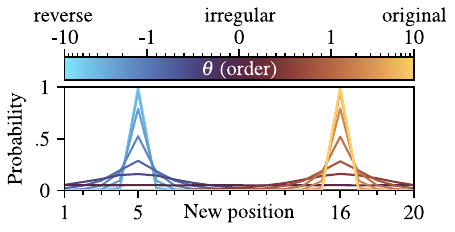}
    \caption{We create a spectrum of synthetic language variants by deterministically permuting words within each sentence. For each sentence length, a permutation is sampled from the Mallows permutation model, where the order parameter $\theta$ controls preference for the original word order. 
    As an example, we show the probability distribution of a word originally at position~16 in a $20$-word sentence.}
    \label{fig:mallows_abstract}
\end{figure}

\paragraph{Approach and contributions}
To overcome these limitations in answering our two research questions, we design a controlled cross-lingual perturbation experiment. We create a continuous spectrum of synthetic word-order variants for ten European languages by deterministically shuffling at the word level.
Our approach uses the Mallows permutation model, which provides a single continuous parameter, the \emph{order} $\theta$, that controls the regularity of word order \cite{mallows-1957-nonnull}. This parameter can be interpreted as a preference for the original word order: Large positive values correspond to the original order; small positive values yield local shuffling; at $\theta = 0$, the order is irregular, such that every word order is equally likely; and negative $\theta$ corresponds to aversion to the original order, up to sentence reversal, see \cref{fig:mallows_abstract}. Crucially, by deterministically shuffling whole words rather than subwords, our method preserves the model-independent global text entropy, vocabulary, and morphology of the original sentences, ensuring that the language variants differ only in terms of word-order regularity.

Our experiments reveal two key findings:
First, by shuffling at the word level rather than at the subword level, we confirm that language-model surprisal increases with more irregular word order, yet it is largely insensitive to sentence reversal.
Second, categorical word-order typology fails to account for language-specific differences, as word-order flexibility is rather a gradient. Instead, vocabulary statistics---Zipf-based coverage metrics, sentence length, and simple proxies for morphological complexity---explain well how robust a language is to free word order in terms of learnability.


\section{Language learnability}
\label{sec:related-work}

Thousands of natural languages exist worldwide \cite{hammarstrom-etal-2025-glottolog}, displaying a wide variety in structural patterns. Here, we are interested in the way these characteristic features influence how difficult a language is to learn for humans and computational models. This section reviews the relation of typological variation to learnability.

\subsection{Language variation}

Natural languages evolve under cognitive and communicative constraints shared by all humans, including limits on information density, redundancy, and processing load \cite{zipf-1935-psychobiology,piantadosi-etal-2012-communicative,hawkins-2014-crosslinguistic,hahn-xu-2022-crosslinguistic,futrell-etal-2020-lossycontext,clark-etal-2023-crosslinguistic}. Within the vast space of symbolic communication systems, natural languages form a small subset shaped by typological correlations \cite{greenberg-1990-universals}. Yet, despite these common forces, they exhibit a striking structural variety.

One prominent example of variation is \emph{word-order flexibility}. The order of subject~(S), verb~(V), and object~(O) is far from uniform across languages: Although the orders SVO or SOV dominate globally \cite{dryer-haspelmath-2024-world}, many languages---such as those from the Slavic and Uralic families---permit comparatively free constituent order, relying on fusional or agglutinative morphology to encode syntactic relations \cite{ponti-etal-2019-modeling,liu-etal-2025-complexity,nijs-etal-2025-word,svenonius-2025-word}. Certain registers, e.g., poetic Latin, even allow nearly unconstrained word order \cite{sampson-2009-linguistic}. 

Word-order structure has been linked to principles such as entropy minimization \cite{franco-sanchez-etal-2024-swap} or uniform information density \cite{clark-etal-2023-crosslinguistic}. Because many factors, especially complex morphology, are intricately connected with word-order flexibility, our goal is to disentangle their contributions to language learnability.

\subsection{Computational learnability} 

Not all languages are equally complex \cite{sampson-etal-2009-language,hahn-etal-2020-universals,koplenig-etal-2023-large}, but it remains unclear whether language models can learn all languages equally well---be they artificial or natural---or whether current architectures systematically favor certain linguistic features \cite{shani-etal-2026-roots}.

In this article, we focus on computational learnability, i.e., how well a model captures the probability distribution of a language, rather than human learnability. Language models are useful in this context because they offer controlled, large-scale experimental setups impossible with human subjects for testing linguistic hypotheses \cite{piantadosi-2024-modern,futrell-mahowald-2025-how}.

Empirical studies indicate that languages differ in how easily models acquire them. Complex inflectional morphology might make languages more difficult \cite{cotterell-etal-2018-are}, although subsequent work found simpler statistics like vocabulary size to be more predictive of model performance than linguistic factors \cite{mielke-etal-2019-what}. 

Tokenization adds another layer of complexity. A performance gap between agglutinative and fusional languages appears to be driven by encoding efficiency rather than morphology itself \cite{arnett-bergen-2025-why}. However, the tokenization properties---including productivity, idiosyncrasy, and fertility---are in turn closely tied to morphology \cite{gutierrez-vasques-etal-2023-languages}. As a result, isolating morphological effects on learnability is difficult, especially since many typological features are more accurately described as gradients than as discrete classes \cite{levshina-etal-2023-why,baylor-etal-2024-multilingual,poelman-etal-2025-confounding}.


\section{Methodology}

One way to disentangle typology and learnability is to create \emph{synthetic language variants} that systematically alter a single typological dimension while keeping the others intact. When based on natural languages, these variants preserve the complexity and irregularity of the original languages but allow targeted manipulations. This idea has been used to explore phenomena like non-concatenative morphology \cite{haley-wilson-2021-deep} or how word order influences translation difficulty \cite{bisazza-etal-2021-difficulty}. We build on this approach by generating synthetic word-order variants to isolate how word order and vocabulary shape learnability.

\subsection{Synthetic word order}
Experiments using word-order perturbations probe how sequence structure affects language models.
Subword-level shuffling has shown that perturbations harm transformer performance, e.g., \citet{kallini-etal-2024-mission}. However, random permutations increase the entropy of the text, complicating the interpretation of model surprisal. Deterministic permutations avoid raising the model-independent entropy by using fixed permutations for each sentence length \cite{clark-etal-2023-crosslinguistic,someya-etal-2025-information}.

Recent cross-lingual studies of computational learnability have arrived at mixed results. \citet{ziv-etal-2025-biasless} found no consistent preference for natural over artificial languages, while \citet{yang-etal-2025-anything} report a moderate inductive bias in favor of natural languages but invariance to violations of certain typological correlations. In contrast, targeted manipulations of specific typological correlations indicate a weak learning bias against those variants \cite{xu-etal-2025-can,el-naggar-etal-2025-which}. Masked language models, in particular, appear largely invariant to shuffling when trained for language-understanding tasks in which word order is partially redundant \cite{hessel-schofield-2021-how, pham-etal-2021-out, papadimitriou-etal-2022-when}.

These studies highlight the value of word-order perturbation for probing learnability, but several limitations recur. First, \emph{perturbations of subwords} split words at inconsistent places that are not linguistically meaningful, which alters both syntax and morphology simultaneously \cite{beinborn-pinter-2023-analyzing,dimarco-fraser-2024-subword}. Second, \emph{disparate shuffling methods} rely on discrete parameters, hindering comparability and control over the degree of perturbation. Third, \emph{narrow language selections}, typically restricted to English, leave cross-lingual variation underexplored.

\subsection{Deterministic shuffling}

We address these limitations of prior perturbation studies through deterministic \emph{word-level shuffling} with a single continuous order parameter $\theta$, applied to a multilingual parallel corpus. This design preserves morphology and keeps  model-independent global entropy practically constant, enabling systematic study of how vocabulary and word-order typology interact in determining learnability.

\paragraph{Our approach}

Intuitively, the desired control parameter \emph{order}~$\theta$ encodes a preference for the original word order of a given sentence in the corpus. By varying $\theta$, we cover the whole spectrum of word-order regularity, ranging from the original order ($\theta \to \infty$), through locally shuffled ($\theta > 0$), to completely irregular ($\theta = 0$), to local shuffling of the reverse order ($\theta < 0$) and full sentence reversal ($\theta \to -\infty$), see \cref{fig:mallows_abstract}.

For example, the sentence \emph{the robot paints the cat} has five words, so we denote the original order as $\pi_0 = (1,2,3,4,5)$. At $\theta = 1$, we might sample a locally shuffled permutation $\pi = (2,1,3,5,4)$ corresponding to the sentence \emph{robot the paints cat the}. At $\theta = 0$, all permutations $\pi$ are equally likely. At $\theta = -7$, the sequence most likely reverses to $\pi = (5,4,3,2,1)$, i.e., \emph{cat the paints robot the}. 

\paragraph{Formal model} We use the \emph{Mallows~$\phi$ model} \cite{mallows-1957-nonnull}, which offers exactly the desired parameter, as the key element of our design. The Mallows model assigns the probability of a permutation $\pi\in \mathfrak{S}_n$ based on the distance $d$ from the original word order $\pi_0 = (1,2,\dotsc,n)$ as
\begin{equation}
    \Pr_{\theta,\pi_0,d}(\pi) = \frac{1}{Z(\theta,d)} \ee^{-\theta d(\pi,\pi_0)}
\label{eq:mallows}
\end{equation}
with the order parameter~$\theta$ and a normalization~$Z$ \cite{crispino-etal-2023-efficient}. Here, the distance metric~$d$ is Kendall's~$\tau$ \cite{kendall-1938-new,tang-2019-mallows}, which counts the minimum number of adjacent swaps to restore the original order~$\pi_0$ from the permutation~$\pi$. With Kendall's~$\tau$, the probability distribution \eqref{eq:mallows} is easy to sample from \cite{fligner-verducci-1986-distance} and yields local shuffling for large~$|\theta|$. Technical details of the Mallows model and an efficient sampling algorithm are given in \cref{sec:mallows}.

\paragraph{Implementation} For each sentence length $n=1,\dotsc,80$ in the corpus of a given language, we sample a single permutation~$\pi^{(n)}$ from the Mallows model and apply it to all sentences of that length. This makes the transformation deterministic, ensuring that the minimum description length (or, equivalently, the model-independent entropy) of the text increases only marginally\footnote{The model-\emph{dependent} entropy for a left-to-right prediction objective may still be sensitive to this nonlocal component, since it cannot know the sentence length in advance.} due to the additional information contained in the $n$ permutations \cite{clark-etal-2023-crosslinguistic,someya-etal-2025-information}.

\section{Experimental setup}
\label{sec:methods}

For our experiments, we generate variants of natural languages with perturbed word order and train identical language models from scratch on each variant. This section outlines the training corpus and languages, pre-processing, shuffling algorithm, model, and evaluation metrics.

\subsection{Data}

\paragraph{Corpus} 
We require a parallel training corpus that encompasses multiple languages with different typology, high quality, and uniform register from multiple speakers, ideally with sentences long enough for word order to play a significant role. The Europarl corpus of European parliamentary speeches meets these criteria \cite{koehn-2005-europarl}.

\paragraph{Language selection}
For interpretability and computational feasibility, we focus on ten out of the 21 languages in Europarl: five languages typically classified as fixed-word-order and five as free-word-order, ensuring variation across morphological type (analytic, fusional, agglutinative). Note that typological categories, including word-order flexibility, are often more aptly described as gradients rather than discrete classes \cite{levshina-etal-2023-why,baylor-etal-2024-multilingual}.
Our sample comprises French, Portuguese (Romance); English, Swedish, Danish (Germanic); Latvian (Baltic); Czech (Slavic); Hungarian, Estonian, Finnish (Finno-Ugric), see \cref{sec:app:languages} for details.

\paragraph{Data preparation}

The definition of a word is convoluted \cite{haspelmath-2023-defining}. We define a word pragmatically as an orthographic unit (whitespace-delimited) to preserve morphological integrity.

Preprocessing involves lowercasing all words and removing punctuation to eliminate positional cues from brackets, commas, quotation marks, etc., see \cref{sec:app:preprocessing}. For each language, we remove sentences longer than 80 words, and then split the text into training, validation, and test sets of \num{650000}, \num{5000}, and \num{5000} sentences, respectively.

\subsection{Model}

We train a lightweight autoregressive language model from scratch with the \textsc{PicoLM} framework \cite{diehlmartinez-etal-2025-pico}, a transformer architecture similar to \textsc{LLaMA} models designed for reproducible research with small language models. The data are tokenized using ByteLevel-BPE, trained from scratch separately for each language with vocabulary size $|V| = 16000$, unless varied. All hyperparameters are listed in \cref{sec:app:hyperparameters}. 

\begin{figure*}[htb]
    \centering
    \includegraphics[width=\linewidth]{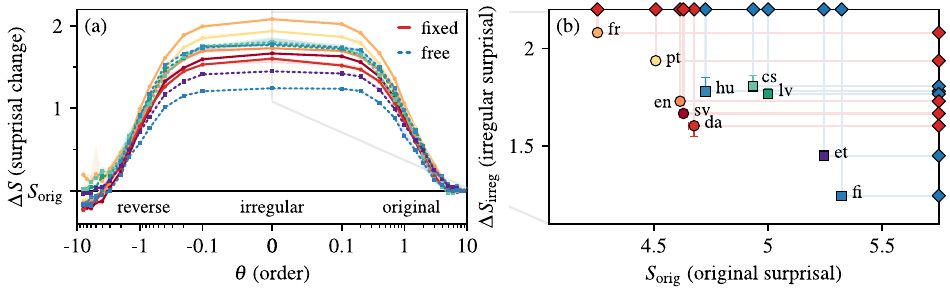}
    \caption{(a) Surprisal change $\Delta S$ due to word-order perturbations with order $\theta$ for each language (named in panel~b). Color shades encode word order: fixed as solid red and free as dashed blue. (b) Zoom-in of surprisal change $\Delta S_\mathrm{irreg}$ at irregular order $\theta = 0$ against the original surprisal $S_\mathrm{orig}$. Red and blue markers are projections onto the axes that indicate a separation of free- and fixed-word-order languages in $S_\mathrm{orig}$ but an overlap in $\Delta S_\mathrm{irreg}$. Transparent bands in panel~(a) and error bars in~(b) show the 25th to 75th percentile over seeds; the lines and points are the median seed, respectively.}
    \label{fig:surprisal}
\end{figure*}

\subsection{Evaluation}

We quantify the learnability of a synthetic language variant with order~$\theta$ via the \emph{model surprisal} $S$, which has been shown to predict human processing difficulty in text parsing \cite{hale-2001-probabilistic,levy-2008-expectationbased,smith-levy-2013-effect}. Surprisal measures how unexpected the observed next subword $w_i$ is to the model given the preceding context $w_{<i}$, i.e., the average negative log-probability
\begin{equation}
    S(\theta) = \frac{1}{N} \sum_{i=1}^{N}- \log\!\big(p_\theta(w_i \mid w_{<i})\big),
\end{equation}
where $p_\theta$ is the model's predictive distribution and $N$ is the total number of tokens in the sequence.\footnote{For the full corpus, we calculate surprisal per subword token over non-overlapping batches due to finite context size.} Each model is evaluated on a test set of the same language variant of order $\theta$ it was trained on.

From an information-theoretic perspective, surprisal is closely related to \emph{entropy}---the average over the Shannon information content (or surprisal) of each single outcome \cite{shannon-1948-mathematical,mackay-2019-information}.
Entropy, and by extension the average surprisal, thus characterize compressibility \cite{schurmann-grassberger-1996-entropy}: \emph{Lower surprisal} means that the model has captured more of the sequence structure, reflecting \emph{greater learnability} of that language variant. Since our shuffling method leaves global entropy essentially constant, any change in surprisal $S(\theta)$ relative to the original surprisal $S_\mathrm{orig} = S(\theta \to \infty)$ of each language, $\Delta S(\theta) = S(\theta) - S_\mathrm{orig}$, is due to the model's sensitivity to the word-order perturbations.


\section{Word-order robustness}
\label{sec:results}

We now turn to how model surprisal varies across language variants for different orders $\theta$, which governs the preference of a given language variant for the original word order. Higher learnability means lower surprisal change $\Delta S$ relative to the unperturbed baseline~$S_\mathrm{orig}$.

\paragraph{Surprisal sensitivity}
First, we observe in \cref{fig:surprisal}~(a) that, across languages, the model surprisal increases with more irregular word order. The surprisal change $\Delta S$ is largest around the fully irregular word order ($\theta = 0$).

Furthermore, sentence reversal ($\theta < 0$) yields almost the same surprisal as the corresponding positive order perturbations ($\theta > 0$).\footnote{The magnitude of this effect is minor with a median surprisal asymmetry~$\Delta S^{+/-}$ of 0.096 across $\theta$, i.e., about $6\,\%$ of the surprisal change due to irregular word order. However, a Wilcoxon signed-rank test on paired differences $\Delta S^{+/-} = \Delta S^+ - \Delta S^-$, aggregated per language, reveals a significant small asymmetry ($p = 0.0098$).} This reflects the symmetry in $\theta$ of the Mallows model \cite{fligner-verducci-1986-distance}, which is largely preserved by the model surprisal, indicating that the models are not strongly sensitive to the typological correlations violated by reversal.

\paragraph{Cross-lingual differences} 
Beyond the overall sensitivity to word-order perturbations observed above, the robustness to shuffling differs by language. Languages allowing freer word order (blue) substantially overlap in~$\Delta S$ with languages that clearly prefer fixed word order (red), suggesting the former are, as a group, no more robust to perturbations, see \cref{fig:surprisal}~(a).

The distinction by free versus fixed word order alone is indeed insufficient: In panel (b), the two groups are clearly separated in baseline surprisal $S_\mathrm{orig}$, yet they overlap in irregular surprisal $\Delta S_\mathrm{irreg} := \Delta S(\theta = 0)$. A Wilcoxon--Mann--Whitney test of the binary word-order flexibility on $\Delta S_\mathrm{irreg}$ yields no significant difference between the groups at irregular word order ($p = 0.55$). Only the extremes---Romance (French, Portuguese) and Finnic (Finnish, Estonian)---are distinguished by both measures. This overlap motivates a search for the driving factors of cross-lingual variation beyond such a categorical word-order flexibility.

\paragraph{Word- vs.\ subword-level shuffling}
\label{sec:subword_shuffling}

\begin{figure}[tbh]
    \centering
    \includegraphics[width=\linewidth]{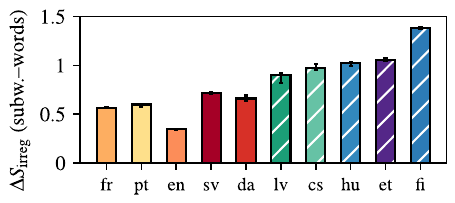}
    \caption{Surprisal difference between word- and subword-level shuffling at irregular word order, with median seed and interquartile ranges.}
    \label{fig:surprisal_difference_word_subword}
\end{figure}

In order to assess whether the shuffling granularity affects cross-lingual learnability, we contrast word- and subword-level shuffling at irregular word order ($\theta = 0$), see \cref{fig:surprisal_difference_word_subword}. Subword-level shuffling yields higher surprisal overall, with markedly larger increases for the Balto-Slavic and Uralic languages, which tend to have more subwords per word. This pattern indicates that breaking morphological integrity can skew cross-lingual comparisons in shuffling experiments and that language-specific vocabulary and tokenization properties related to morphological complexity play a central role in shaping robustness to word-order perturbations.

\section{The role of the vocabulary}

Zipfian distributions \cite{piantadosi-2014-zipfs} relate closely to morphological complexity and word order \cite{liu-etal-2025-complexity}. \emph{Vocabulary structure}---in the sense of word and subword frequencies, the relation of subwords to words, and sequence length---thus varies systematically between languages and characterizes a language beyond word order. Our aim is to derive latent structures from a set of simple metrics of vocabulary structure that explain cross-lingual variation in robustness to free word order.

\subsection{Vocabulary metrics} 

\begin{figure}[tb]
    \centering
    \includegraphics[width=\linewidth]{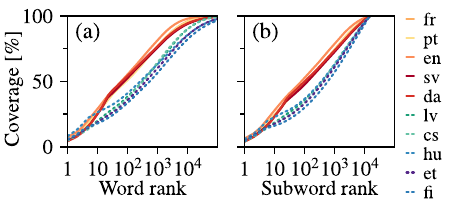}
    \caption{Percentage of (a) words and (b) subwords in the corpus accounted for by the most frequent vocabulary items. This coverage increases more slowly for languages with freer word order compared to languages with relatively fixed word order (shades of blue and red, respectively).}
    \label{fig:coverage}
\end{figure}

One aspect of vocabulary structure is coverage~$C(r)$: the proportion of the corpus accounted for by the $r$ most frequent word or subword types. Coverage is the cumulative sum of the rank-frequency distribution described by Zipf's law \cite{zipf-1949-human}.

In \cref{fig:coverage}~(a), word coverage clearly clusters languages into free word order (blues) and fixed word order (reds). Subword coverage in panel~(b) preserves this separation after tokenization. This intrinsic typological grouping, which also captures the effect of tokenization on the vocabulary, renders vocabulary structure a strong candidate for explaining cross-lingual variation in surprisal.

This clustering suggests that coverage offers a more informative basis for predicting and thus explaining cross-lingual surprisal than a binary free/fixed typology. To capture the essence of the coverage curves, we select four characteristics: word and subword coverage at rank $100$, the integral of word coverage, and the similarity between word and subword coverage, defined in \cref{sec:app:coverage_metrics}. We complement this predictor set with other simple metrics of vocabulary structure: sentence length (average words and subwords) and proxies for morphological complexity (fertility, average word length, number of unique word types), see \cref{sec:app:pls_predictors}.

\subsection{Explaining word-order robustness}

To identify latent structure in the predictors and assess their explanatory power for language-specific surprisal, we employ multivariate partial-least-squares (PLS) regression. PLS is well-suited for this setting of highly collinear predictors\footnote{See the correlation matrix in \cref{sec:app:correlation-matrix}.} and a small sample size ($n = 10$ languages) with multivariate responses ($S(\theta)$ at 28 values of $\theta$ per language). PLS accomplishes dimensional reduction by creating latent components while retaining predictive power to explain the variance between predictors and response variables. Leave-one-language-out cross-validation identifies two components as the optimal number, with an overall predictive performance of $R^2 = 0.97$ explained variance, see \cref{fig:latents_components}~(a) and (b) in \cref{sec:app:loadings} for details.

\begin{figure}[tb]
    \centering
    \includegraphics[width=\linewidth]{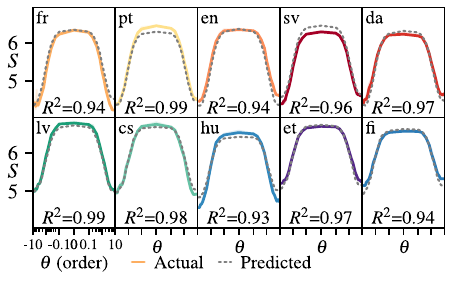}
    \caption{The absolute surprisal $S(\theta) = S_\mathrm{orig} + \Delta S(\theta)$ per language modeled through a set of vocabulary statistics, encompassing coverage, sentence length, and proxies for morphological complexity. The predictions are cross-validated through leave-one-language-out: Each language is predicted solely on the basis of its own vocabulary statistics by a model trained on the surprisal of the other languages and their predictors.}
    \label{fig:latents_per_language}
\end{figure}

Predictions from the cross-validated models capture the curve $S(\theta)$ closely and explain most of the variance per left-out language, ranging from $R^2 = 0.93$ for Hungarian to $R^2 = 0.99$ for Portuguese and Latvian, see \cref{fig:latents_per_language}.

\begin{figure}[tb]
    \centering
    \includegraphics[width=\linewidth]{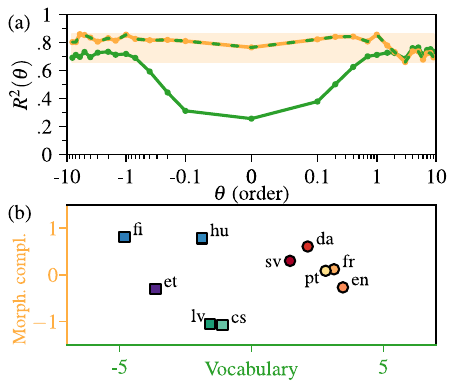}
    \caption{(a) The cross-validated explained variance per slice of $\theta$ of only the vocabulary component (green) with mean $\bar{R}^2 = 0.65$, ranging from $0.26$ to $0.76$, and of both components (green-yellow) with mean $\bar{R}^2 = 0.79$, ranging from $0.66$ to $0.86$. (b) PLS scores of the main component (vocabulary) and the secondary component (morphological complexity).}
    \label{fig:latents_scores_and_r2theta}
\end{figure}

Variance explained per slice of $\theta$ ranges from $R^2 = 0.66$ to $0.86$ with mean $\bar{R}^2 = 0.79$, see \cref{fig:latents_scores_and_r2theta}~(a), demonstrating that the predictions are stable across various forms of word-order perturbations. The first component (vocabulary) on its own explains the original and reverse order with $\bar{R}^2 = 0.65$, ranging from $0.26$ to $0.76$. The second component (complex morphology: unique word types and word length) is therefore necessary to explain the regime of irregular word order.

\Cref{fig:latents_scores_and_r2theta}~(b) shows the learned latent structure. The primary component comprises coverage, and to a lesser extent sentence length and morphological complexity and structurally aligns equally across all $\theta$. The secondary component reflects morphological complexity and is most associated with irregular order perturbations at small order $|\theta|$, see \cref{fig:latents_components}~(b) in \cref{sec:app:loadings}. 

In summary, the vocabulary metrics explain surprisal $S(\theta)$ across languages and perturbation orders $\theta$ better than the binary free/fixed-word-order typology. While coverage explains the original- and reverse-order surprisal, complex morphology is a main factor in what makes a language more robust to shuffling.

\subsection{Vocabulary size}

\begin{figure*}[htb]
    \centering
    \includegraphics[width=\linewidth]{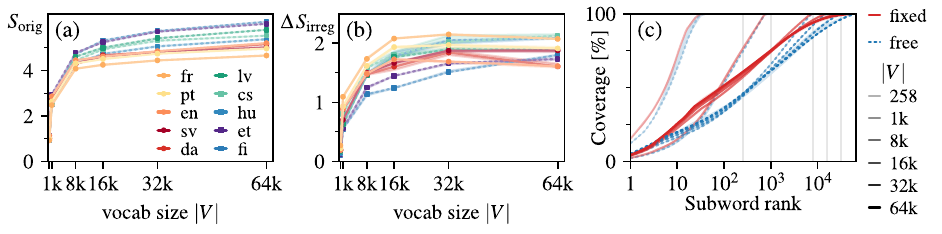}
    \caption{Effect of the vocabulary size on (a) the original surprisal $S_\mathrm{orig}$ and (b) the surprisal change at irregular word order $\Delta S_\mathrm{irreg}$. (c) Subword coverage per vocabulary size, grouped by free and fixed word order, where line transparency and the vertical gray lines encode the vocabulary size. Colored lines show medians and shaded regions show interquartile ranges, computed across random seeds for panels (a) and (b) and across languages for panel (c).}
    \label{fig:vocab_size}
\end{figure*}

Tokenization compresses the word vocabulary into a subword vocabulary and may influence cross-lingual differences in word-order robustness. We examine this by varying the vocabulary size~$|V|$ for the original ($\theta = \infty$) and irregular word-order ($\theta = 0$) condition, see \cref{fig:vocab_size}.

The original surprisal~$S_\mathrm{orig}$ begins to separate the free- and fixed-word-order languages above $|V| = 8000$ and the ordering of the languages remains largely consistent at larger vocab sizes, see panel~(a). Conversely, the surprisal change at irregular word order~$\Delta S_\mathrm{irreg}$ in panel~(b) converges between the languages at larger vocabulary sizes. This overlap stems from a downward trend or plateau of languages with rather fixed word order, while the other languages keep increasing up to $|V| = \num{64000}$. 

Panel~(c) shows that this separation in $S_\mathrm{orig}$ coincides with two clusters emerging in subword coverage above $|V| = 8000$. Free-word-order languages make greater use of low-frequency subwords, resulting in a more slowly increasing coverage. This more fine-grained representation of the binary free-vs.-fixed classification through coverage is consistent with the observation from \cref{sec:app:loadings} that the vocabulary PLS component, which primarily explains the original surprisal, is strongly associated with coverage. These results highlight how tokenization modulates crosslingual differences in word-order learnability.


\section{Discussion}
\label{sec:discussion}

Our experiments show that higher word-order irregularity hinders language-model learning across languages, but models remain largely insensitive to reversal. Cross-lingual variation is better predicted by vocabulary structure than by binary word-order flexibility. Our results clarify the factors that contribute to computational word-order learnability.

\paragraph{Relation to prior work}
The sensitivity to irregular word order reflects a \emph{locality bias} \cite{choshen-abend-2019-automatically}, and extends recent work on artificial-language learnability \cite{kallini-etal-2024-mission,xu-etal-2025-can,yang-etal-2025-anything,kallini-potts-2025-language,el-naggar-etal-2025-which} to controlled word-level shuffling. By using a unified perturbation spectrum, our approach preserves morphological integrity and avoids confounds of disparate shuffling schemes in earlier work. Furthermore, our findings challenge claims that language models can learn all languages alike \cite{chomsky-2023-noam,katzir-2023-why,leivada-etal-2025-large,ziv-etal-2025-biasless}.

Previous studies on sentence reversal found small and inconsistent differences in surprisal \cite{yang-etal-2025-anything,ziv-etal-2025-biasless}. In contrast, our broader analysis across the continuous order spectrum exhibits near symmetry with respect to reversal, slightly favoring reverse variants, matching a bias towards head-initial synthetic grammars \citet{el-naggar-etal-2025-gcgbased}.

For \emph{cross-lingual differences}, our results show, in line with \citet{mielke-etal-2019-what}, that simpler vocabulary statistics and sentence length suffice as predictors, whereas \citet{cotterell-etal-2018-are} emphasize the role of complex morphology. The apparent distinction between typology or simpler statistics as an explanation \cite{yang-etal-2025-anything,arnett-bergen-2025-why} is resolved if the vocabulary-based measures quantify aspects of typology---such as word-order robustness---more richly than coarse labels \cite{levshina-etal-2023-why,baylor-etal-2024-multilingual}. Although previous work suggests that word order is causally responsive to morphological complexity \cite{nijs-etal-2025-word}, our PLS analysis captures only associations between vocabulary structure and word-order flexibility and remains agnostic to their causal relationship in language evolution.

\paragraph{Architecture and mechanisms}
Several architectural factors may play into these results.
First, the prediction objective of autoregressive transformer models limits the context for predicting the next token to previous tokens. Interestingly, since our deterministic shuffling can, in principle, be reversed if the sentence length is known, this shuffling introduces a nonlocal component to the language because the model does not know the exact sentence length ahead of time. Thus, the autoregressive nature of the models may underlie the general sensitivity to word-order perturbations across languages. In alignment with prior work, masked language models might therefore be less sensitive to irregular word order \cite{hessel-schofield-2021-how, pham-etal-2021-out, papadimitriou-etal-2022-when}. 

Second, larger vocabularies tend to reduce the irregular surprisal change for a subset of languages in our experiments. At large vocabulary sizes, the embedding parameters begin to outnumber the core model parameters. Possibly, languages that rely more on rare subwords and have high type diversity (see \cref{sec:app:pls_predictors}) may be disadvantaged by limited model capacity and slower convergence of low-frequency vocabulary items in the embedding space during training \citep{tao-etal-2024-scaling,papadimitriou-prince-2025-vocabulary}. 

Positional encoding also affect how shuffled input is represented, with distinct correlation patterns for words and subwords \cite{abdou-etal-2022-word}. Future work should disentangle the architectural features of prediction objective, tokenization, and positional encoding.


\section{Conclusion}
\label{sec:conclusion}

We set out to understand what makes a language computationally difficult to learn for language models, using a spectrum of synthetic language variants with perturbed word order. Our experiments reveal three main findings:
(1) Irregular word order decreases computational learnability;
(2) but language models are largely insensitive to subtler violations of typological correlations introduced by sentence reversal; and 
(3) the robustness of a language to word-order perturbations is predicted better by vocabulary structure (Zipf-based coverage, sentence length, and morphological complexity) than by the coarse distinction into free and fixed word order. Morphological complexity proxies are most relevant for explaining robustness against strongly irregular word order.  

These findings establish that simple vocabulary metrics constitute a powerful basis for explaining cross-lingual differences in word-order learnability, providing a more comprehensive predictor than categorical typological classifications. Vocabulary structure is an integral part of interpreting model surprisal in shuffling experiments. The robustness to word-order perturbations is thus an inherent property of natural languages that gives insight into the interplay of vocabulary and word order.

Future work should examine how model design---such as masking, tokenization, and positional encoding---modulates sensitivity to word-order perturbations and compare models with human behavior to assess cognitive plausibility. We also foresee extending our method of word-level shuffling to morpheme-level and part-of-speech-dependent perturbations. Such research linking language features and model architecture advances the understanding of language learnability.

\section*{Limitations}

The present study should be interpreted in light of several limitations.

\paragraph{Corpus}

We use a single high-quality parallel corpus to ensure comparability across languages, yet it is limited to 21 European languages, of which we selected ten for a focused analysis and computational feasibility. Extending to more corpora would allow for a more diverse set of language typologies to be included \cite{ploeger-etal-2024-what,ploeger-etal-2025-principled} at the cost of more noise and heterogeneity in the data.

\paragraph{Human learnability} We use language models as a tool to study learnability, yet the learnability of a language model does not necessarily generalize to humans. Comparisons with human data, e.g., eye-tracking studies \cite{schad-etal-2010-eye}, could help evaluate cognitive plausibility.

\paragraph{Model size} Since our experimental setup requires a large number of models to be trained, the model size is limited in order to achieve reasonable training times. This trade-off could impact vocabulary size effects at very large vocabularies, for which embedding parameters dominate.

\paragraph{Typology} We group languages into \enquote{free} and \enquote{fixed} word order, but typology is a gradient \cite{levshina-etal-2023-why,baylor-etal-2024-multilingual}. A comparative analysis of other continuous typological measures, e.g., subject-object-order entropy, with the vocabulary-structure measures we describe here may yield a more nuanced understanding. 

\paragraph{Evaluation} We evaluate the global surprisal on a test set. An interesting extension would be to assess whether all tokens contribute uniformly or whether surprisal effects can be attributed to breaking certain language-specific collocations, e.g., determiner-adjective-noun constructions.

\section*{Ethical considerations}

\paragraph{Synthetic languages}

Our study uses synthetic languages (also called \enquote{artificial languages}). There is a wide spectrum of languages, ranging from formal and highly unnatural to attested languages. It is important not to conflate different categories on this spectrum. In our study, we focus on languages that are perturbed only on the dimension of word order, while maintaining the complexity of natural language in terms of lexicon and morphology.

\paragraph{Environmental impact}

Training models, even with comparatively few parameters, leads to computational cost and CO$_2$ emissions. We encourage future work to consciously evaluate the need for large-scale studies in order to curtail the ever-increasing environmental impact of our information infrastructure.

Development and training models for this study required approximately $150\,\text{kcore-hours}$. The models were trained on one A100 GPU with $\SI{40}{\giga\byte}$ with $1400$ models for the scan in $\theta$ and $600$ models for the scan in vocabulary size~$|V|$.

\section*{Acknowledgments}

We thank the anonymous reviewers for their thoughtful feedback.
J.J.'s research is supported by a Dutch National Science Organisation (NWO) grant (VI.Vidi.221C.009).
L.B.'s research is partially supported by an Impulsprofessur grant from the zukunft.niedersachsen program and by a VENI grant (Vl.Veni.211C.039) from the NWO. 
The authors gratefully acknowledge computing time
provided to them at the GWDG HPC cluster.

\appendix

\begin{table*}[htbp]
    \small
	\centering
	\caption{Language selection with name and ISO-code abbreviation, grouped by branch and family (IE: Indo-European). We list word-order flexibility and morphology.}
    \label{tab:languages}
	\begin{tabular}{l  c  l  c   *2c l}
		\toprule
		Language & ISO & Branch & Family & Flexibility & Morphology & Reference \\
		\midrule
		English 	& en & \multirow{3}{*}{Germanic} & \parbox[t]{2mm}{\multirow{5}{*}{IE}} & \multirow{5}{*}{fixed} & analytic & \cite{aarts-2011-oxford} \\
		Danish 		& da & & & & analytic & \cite{lundskaer-nielsen-holmes-2015-danish} \\
		Swedish 	& sv & & & & analytic & \cite{holmes-hinchliffe-2013-swedish} \\
		Portuguese 	& pt & \multirow{2}{*}{Romance} & & & fusional & \cite{kabatek-2022-manual,harris-vincent-2012-romance} \\
		French 		& fr & & & & fusional & \cite{harris-vincent-2012-romance} \\
		\midrule
		Latvian 	& lv & Baltic & IE & \multirow{5}{*}{free} & fusional & \cite{praulins-2012-latvian} \\
		Czech 		& cs & Slavic & IE & & fusional & \cite{naughton-2008-czech} \\
		Hungarian 	& hu & Ugric & \parbox[t]{2mm}{\multirow{3}{*}{\rotatebox[origin=c]{90}{Uralic}}} & & agglutinative & \cite{kenesei-etal-2002-hungarian} \\
		Estonian 	& et & \multirow{2}{*}{Finnic} & & & agglutinative & \cite{harms-1997-estonian} \\
		Finnish 	& fi & & & & agglutinative & \cite{sulkala-karjalainen-2012-finnish,karlsson-2017-finnish} \\
	\end{tabular}
\end{table*}

\section{Shuffling algorithm} 
\label{sec:mallows}

The idea of sampling one permutation~$\pi$ for each sentence length~$n$ to shuffle a language corpus deterministically has been used in \cite{someya-etal-2025-information}. We introduce the Mallows model \citep{mallows-1957-nonnull,tang-2019-mallows} as a unifying probabilistic method for selecting the permutation~$\pi$ for each sentence length~$n$.

The Mallows model assigns the probability of a permutation $\pi\in \mathfrak{S}_n$, based on the original word order $\pi_0 = (1,2,\dotsc,n)$, as
\begin{equation}
    \Pr_{\theta,\pi_0,d}(\pi) = \frac{1}{Z(\theta,d)} \ee^{-\theta d(\pi,\pi_0)}
\end{equation}
where the control parameter $\theta$ is the order (also called dispersion or concentration \cite{crispino-etal-2023-efficient} and analogous to an inverse temperature $\beta$), $d$~is a distance metric measuring the discrepancy between $\pi$ and $\pi_0$, and $Z$ is the partition function that normalizes the distribution. The order $\theta$ is interpreted as how preferred the original order $\pi_0$ is by the probability distribution.

Since the Mallows $\phi$ model \cite{tang-2019-mallows} is easy to sample from (see \citet{fligner-verducci-1986-distance} for details), we use Kendall's $\tau$ as distance metric \cite{kendall-1938-new},
\begin{equation}
    d_\tau(\pi \circ \pi_0^{-1}) = \mathrm{inv}(\pi \circ \pi_0^{-1})\,,
\end{equation}
where $\mathrm{inv}(\pi) := |\{(i,j) \in [n]^2 : i < j \wedge \pi(i) > \pi(j)\}|$, that is, $d_\tau$ is the minimum number of adjacent swaps to restore the central order $\pi_0$ from the permutation $\pi$.

According to \citet{fligner-verducci-1986-distance}, the Mallows $\phi$ model (for permutations $\pi \in \mathfrak{S}_n$ of length $n$) has the mean
\begin{equation}
    \EE_\theta(d_\tau) = \frac{n \ee^{-\theta}}{1 - \ee^{-\theta}} - \sum_{j=1}^{n} \frac{j \ee^{-j\theta}}{1 - \ee^{-j\theta}}
\end{equation}
and variance
\begin{equation}
    \VV_\theta(d_\tau) = \frac{n \ee^{-\theta}}{(1 - \ee^{-\theta})^2} - \sum_{j=1}^{n} \frac{j^2 \ee^{-j\theta}}{(1 - \ee^{-j\theta})^2}
\end{equation}
with maximum distance \cite{kendall-1938-new}
\begin{equation}
    d_{\mathrm{max}} = \binom{n}{2} = \frac{n (n - 1)}{2}
\end{equation}
between permutations and maximum variance \cite[p. 257]{feller-1968-introduction}
\begin{equation}
    v_\mathrm{max} = \frac{n (n-1) (2n + 5)}{72}\,,
\end{equation}
respectively.

\begin{figure}[ht]
    \centering
    \includegraphics[width=1\linewidth]{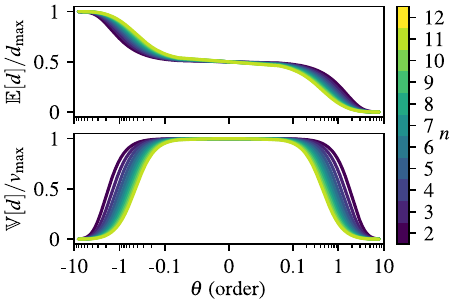}
    \caption{Normalized analytical mean and variance of Kendall's $\tau$ with Mallows shuffling over the order $\theta$ for different sentence lengths $n$.}
    \label{fig:mallows}
\end{figure}

\Cref{fig:mallows} shows the normalized mean and variance of the Mallows $\phi$ distribution by the order $\theta$ for different sentence lengths $n$.

\section{Language selection}
\label{sec:app:languages}

We select ten out of the 21 languages available in Europarl for our experiments: Five languages commonly classified---either categorically or via continuous measures \cite{siewierska-2010-constituent,levshina-etal-2023-why}---as fixed-word-order and five as free-word-order, namely: English, Danish, Swedish (Germanic), French, Portuguese (Romance), Latvian (Baltic), Czech (Slavic), Hungarian, Estonian, Finnish (Finno-Ugric), see \cref{tab:languages}. The dominant or neutral word order of all these languages is SVO \cite{siewierska-2010-constituent}.

\newpage

\section{Preprocessing}
\label{sec:app:preprocessing}

We cleaned the raw Europarl sentences prior to shuffling using these steps.

\begin{itemize}
    \item Remove empty or punctuation-only sentences, speaker and language labels, and obvious non-speech content.
    \item Fix or remove Unicode artifacts: replace soft hyphens (U+00AD) with \enquote{-}; remove replacement characters (U+FFFD) and zero-width spaces (U+200B); drop lines containing URLs.
    \item Strip nested parenthetical and bracketed content, quotation marks, and apostrophes while preserving enclosed text.
    \item Normalize punctuation: remove stray commas, split at semicolons and colons outside words; clean bullets and dashes, replacing with hyphens where appropriate.
    \item Collapse whitespace, lowercase text, and ensure terminal punctuation.
    \item Apply minimal language-specific rules: remove mistaken spaces in Finnish abbreviations (\enquote{EU: n} $\rightarrow$ \enquote{EU:n}).
\end{itemize}

For each language, we remove sentences longer than 80 words, and then split into training, validation, and test sets of \num{650000}, \num{5000}, and \num{5000} sentences, respectively.

Full preprocessing code and regex rules are available in our\, \href{https://gitlab.gwdg.de/huds/projects/shuffle/-/tree/v1.0.2}{\raisebox{-0.15em}{\includegraphics[height=0.9em]{logo_gitlab.pdf}}\;code repository}.

\newpage
\section{Training parameters}
\label{sec:app:hyperparameters}

\begin{table}[htbp]
    \small
    \caption{Hyperparameters for the \textsc{PicoLM} models used in our experiments.}
    \label{tab:hyperparameters}
    \centering
    \begin{tabular}{l r}
        \toprule
        \textbf{Parameter} & \textbf{Value} \\
        \midrule
        Architecture & Transformer decoder \\
        Total parameters & 50.5\,M\\
        Layers & 12 \\
        Embedding size & 384 \\
        Attention heads & 12 \\
        Attention KV heads & 4 \\
        Hidden dimension & \num{1536} \\
        Sequence length & 512 \\
        Tokenizer & ByteLevel BPE \\
        Tokenizer min. freq. & 2 \\
        Vocabulary size~$|V|$ & \num{16000} / varied\\
        Optimizer & AdamW \\
        Learning rate & 0.0014 \\
        Learning rate schedule & Linear \\
        Warmup steps & 5 \\
        Batch size (training) & 64 \\
        Training steps & \num{1000} \\
        Order $\theta$ & varied / $\{0,9\}$\\
        \bottomrule
    \end{tabular}
\end{table}

\Cref{tab:hyperparameters} lists the hyperparameters of training the language models for our experiments. Each model was trained on one A100 GPU.

We generate five random seeds per language variant and apply the deterministic word-level shuffling for a range of orders $\theta$, training each model with batch size 64 for $\num{1000}$ steps. The vocabulary size is $|V| = \num{16000}$ when varying the order $\theta$, with roughly log-scaled $\theta \in [-9,9]$. When the vocabulary size is varied as $|V| = \{258, \num{1000}, \num{8000}, \num{16000}, \num{32000}, \num{64000}\}$ the order is chosen as $\theta \in \{0,9\}$. Note that $256+2$ corresponds to character-level tokenization (with two additional padding and end-of-text tokens).

\clearpage
\section{PLS regression}
\label{sec:app:pls}

Here, we define and list predictors used for the partial-least-squares (PLS) regression analysis.

\begin{table*}[htbp]
\small
\centering
\caption{Vocabulary statistics for each language: Word and subword coverage at rank 100, coverage similarity and integral; sentence length in subwords and words; morphological complexity measured by fertility, word length in characters, and word types.}
\label{tab:lang_stats}
\begin{tabular}{l | l
                r r r
                r r
                r r
                r r
                r}
\toprule
    &
    & \multicolumn{4}{c}{Coverage} 
    & \multicolumn{2}{c}{Sentence length} 
    & \multicolumn{3}{c}{Morph.\ complexity} \\
\cmidrule(lr){3-6} \cmidrule(lr){7-8} \cmidrule(lr){9-11}
Lang. & Order & $C_{\mathrm{w},100}$ & $C_{\mathrm{s},100}$ & $C$ simil.\ & $\int C_\mathrm{w}$ & Subw./sent. & Words/sent. & Fertility & Word len. & Types \\
\midrule
fr & \multirow{5}{*}{Fixed} & 52.7 & 49.0 & 0.974 & 61.0 & 28.1 & 24.4 & 1.15 & 6.02 & 96\,727 \\
pt &  & 51.0 & 47.9 & 0.979 & 60.6 & 28.1 & 24.2 & 1.16 & 6.03 & 108\,442 \\
en &  & 55.4 & 52.6 & 0.976 & 62.6 & 26.4 & 23.8 & 1.11 & 5.70 & 70\,536 \\
sv &  & 52.6 & 47.5 & 0.971 & 60.1 & 24.4 & 20.4 & 1.20 & 6.22 & 177\,002 \\
da &  & 54.2 & 49.6 & 0.971 & 60.8 & 25.7 & 21.7 & 1.19 & 6.09 & 179\,915 \\ \hline
lv & \multirow{5}{*}{Free}  & 38.3 & 35.7 & 1.006 & 52.9 & 23.1 & 18.2 & 1.27 & 6.88 & 156\,845 \\
cs &   & 37.1 & 34.0 & 1.005 & 52.6 & 24.8 & 19.6 & 1.27 & 6.35 & 169\,003 \\
hu &   & 40.9 & 36.6 & 1.017 & 53.9 & 24.8 & 18.7 & 1.33 & 7.23 & 307\,197 \\
et &   & 35.5 & 33.0 & 1.044 & 50.3 & 22.1 & 16.5 & 1.34 & 7.40 & 283\,165 \\
fi &   & 33.5 & 29.8 & 1.048 & 48.3 & 23.2 & 16.2 & 1.43 & 8.33 & 363\,154 \\
\bottomrule
\end{tabular}
\end{table*}

\subsection{Definition of coverage metrics}
\label{sec:app:coverage_metrics}

\begin{figure}[tbh]
    \centering
    \includegraphics[width=\linewidth]{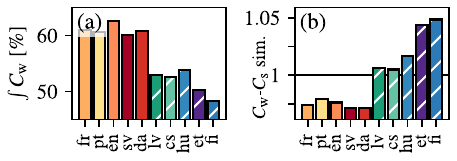}
    \caption{Coverage measures: the (a) word coverage integral and (b) word-subword-coverage similarity.}
    \label{fig:coverage_measures}
\end{figure}

We calculate the coverage integral as the area under the word coverage curve $C_\mathrm{w}$ per log-rank up to $r_\mathrm{max} = 10^5$:
\begin{equation}
   \frac{1}{\log(r_\mathrm{max})} \int_1^{r_\mathrm{max}} C_\mathrm{w} \;\mathrm{d}(\log(r))\,.
\end{equation}

The coverage similarity relates word and subword coverage through a regression slope $m$ in log-space without intercept,
\begin{equation}
    m = \frac{\sum_{r=1}^{r_\mathrm{max}} w_r \, C_\mathrm{w}(r) \, C_\mathrm{s}(r)}{\sum_{r=1}^{r_\mathrm{max}} w_r \, (C_\mathrm{w}(r))^2}\,,
\end{equation}
with weights given by $\log(r)$.

Both the coverage integral and similarity, visualized in \cref{fig:coverage_measures}, clearly separate free- from fixed-word-order languages.

\subsection{Regression factors}
\label{sec:app:pls_predictors}

All vocabulary statistics used as predictors for the PLS analysis are listed by language in \cref{tab:lang_stats} along with the classification into free- and fixed word order. The predictors are grouped by coverage, sentence length, and morphological complexity.

The morphological complexity metrics are: fertility, i.e., the average number of subwords per word; average word length in characters; and number of types in the sense of unique words, i.e., the word vocabulary of the corpus.

\begin{figure}[tbhp]
    \centering
    \includegraphics[width=\linewidth]{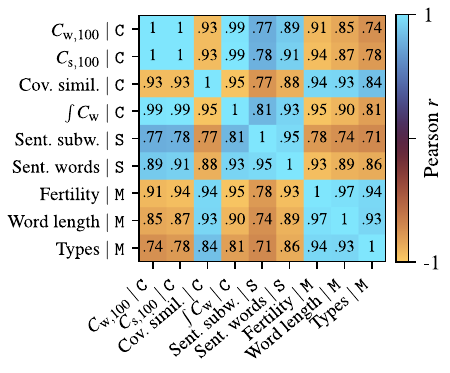}
    \caption{Correlation matrix of the predictors, grouped by coverage (C), sentence length (S), and proxies of morphological complexity (M).}
    \label{fig:correlation-matrix}
\end{figure}

\subsection{Correlation of factors}
\label{sec:app:correlation-matrix}

The correlation matrix in \cref{fig:correlation-matrix} shows that all predictors are highly correlated, motivating the use of a PLS regression.

\subsection{Latent components}
\label{sec:app:loadings}

\begin{figure*}[!tbhp]
    \centering
    \includegraphics[width=\linewidth]{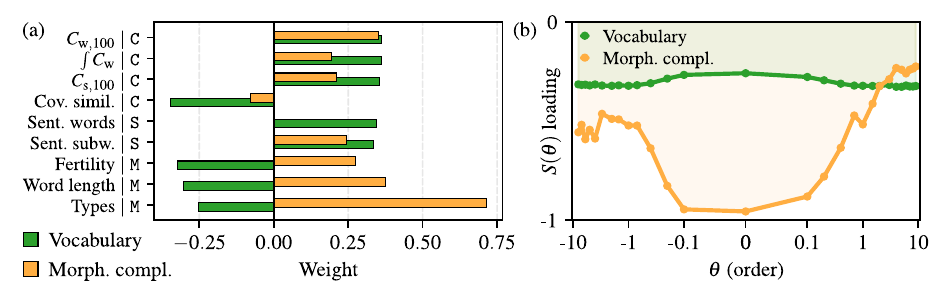}
    \caption{(a) Predictor and (b) response loadings of the vocabulary and morphological complexity component.}
    \label{fig:latents_components}
\end{figure*}

\Cref{fig:latents_components}~(a) shows the two components identified by the PLS regression: The vocabulary component loads on all predictors, but more strongly on coverage; the morphological-complexity component loads primarily on fertility, word length, and word types. 

In panel~(b), we see that the vocabulary component is structurally uniform across the spectrum of~$\theta$, whereas the morphological-complexity component aligns most with the irregular word order at small $|\theta|$.

Predicting per $\theta$ only with the single binary predictor of free vs.\ fixed word order yields $R^2 = 0.44$ to $0.85$ with mean $\bar{R}^2 = 0.65$ and lowest per language $R^2 = 0.77$ for Portuguese and overall $R^2 = 0.94$. A single non-predictive feature like characters per token yields per-$\theta$ mean $\bar{R}^2 = -0.02$ and overall $R^2 = 0.85$. The $R^2$ of the latter remains relatively high because of the shared structure in the $S(\theta)$ curves. We observe that word-order flexibility is not as comprehensive as the combination of vocabulary-structure metrics but explains more than the non-predictive characters per token.


\begin{thebibliography}{83}
\providecommand{\natexlab}[1]{#1}

\bibitem[{Aarts(2011)}]{aarts-2011-oxford}
Bas Aarts. 2011.
\newblock \href
  {https://global.oup.com/academic/product/oxford-modern-english-grammar-9780199533190?cc=de&lang=en&}
  {\emph{Oxford {{Modern English Grammar}}}}, 1st edition.
\newblock Oxford University Press, New York.

\bibitem[{Abdou et~al.(2022)Abdou, Ravishankar, Kulmizev, and
  S{\o}gaard}]{abdou-etal-2022-word}
Mostafa Abdou, Vinit Ravishankar, Artur Kulmizev, and Anders S{\o}gaard. 2022.
\newblock \href {https://doi.org/10.18653/v1/2022.acl-long.476} {Word order
  does matter and shuffled language models know it}.
\newblock In \emph{Proceedings of the 60th {{Annual Meeting}} of the
  {{Association}} for {{Computational Linguistics}} ({{Volume}} 1: {{Long
  Papers}})}, pages 6907--6919, Dublin, Ireland. Association for Computational
  Linguistics.

\bibitem[{Arnett and Bergen(2025)}]{arnett-bergen-2025-why}
Catherine Arnett and Benjamin~K. Bergen. 2025.
\newblock \href {https://aclanthology.org/2025.coling-main.441/} {Why do
  language models perform worse for morphologically complex languages?}
\newblock In \emph{Proceedings of the 31st {{International Conference}} on
  {{Computational Linguistics}}}, pages 6607--6623, Abu Dhabi, UAE. Association
  for Computational Linguistics.

\bibitem[{Baylor et~al.(2024)Baylor, Ploeger, and
  Bjerva}]{baylor-etal-2024-multilingual}
Emi Baylor, Esther Ploeger, and Johannes Bjerva. 2024.
\newblock \href {https://doi.org/10.18653/v1/2024.eacl-short.6} {Multilingual
  gradient word-order typology from universal dependencies}.
\newblock In \emph{Proceedings of the 18th {{Conference}} of the {{European
  Chapter}} of the {{Association}} for {{Computational Linguistics}}
  ({{Volume}} 2: {{Short Papers}})}, pages 42--49, St. Julian's, Malta.
  Association for Computational Linguistics.

\bibitem[{Beinborn and Pinter(2023)}]{beinborn-pinter-2023-analyzing}
Lisa Beinborn and Yuval Pinter. 2023.
\newblock \href {https://doi.org/10.18653/v1/2023.emnlp-main.272} {Analyzing
  cognitive plausibility of subword tokenization}.
\newblock In \emph{Proceedings of the 2023 {{Conference}} on {{Empirical
  Methods}} in {{Natural Language Processing}}}, pages 4478--4486, Singapore.
  Association for Computational Linguistics.

\bibitem[{Bisazza et~al.(2021)Bisazza, {\"U}st{\"u}n, and
  Sportel}]{bisazza-etal-2021-difficulty}
Arianna Bisazza, Ahmet {\"U}st{\"u}n, and Stephan Sportel. 2021.
\newblock \href {https://doi.org/10.1162/tacl_a_00424} {On the difficulty of
  translating free-order case-marking languages}.
\newblock In \emph{Transactions of the {{Association}} for {{Computational
  Linguistics}}}, volume~9, pages 1233--1248, Cambridge. MIT Press.

\bibitem[{Chomsky(2023)}]{chomsky-2023-noam}
Noam Chomsky. 2023.
\newblock \href
  {https://www.nytimes.com/2023/03/08/opinion/noam-chomsky-chatgpt-ai.html}
  {Noam {{Chomsky}}: {{The}} false promise of {{ChatGPT}}}.
\newblock \emph{The New York Times}.

\bibitem[{Choshen and Abend(2019)}]{choshen-abend-2019-automatically}
Leshem Choshen and Omri Abend. 2019.
\newblock \href {https://doi.org/10.18653/v1/K19-1028} {Automatically
  extracting challenge sets for non-local phenomena in neural machine
  translation}.
\newblock In \emph{Proceedings of the 23rd {{Conference}} on {{Computational
  Natural Language Learning}} ({{CoNLL}})}, pages 291--303, Hong Kong, China.
  Association for Computational Linguistics.

\bibitem[{Clark et~al.(2023)Clark, Meister, Pimentel, Hahn, Cotterell, Futrell,
  and Levy}]{clark-etal-2023-crosslinguistic}
Thomas~Hikaru Clark, Clara Meister, Tiago Pimentel, Michael Hahn, Ryan
  Cotterell, Richard Futrell, and Roger Levy. 2023.
\newblock \href {https://doi.org/10.1162/tacl_a_00589} {A cross-linguistic
  pressure for uniform information density in word order}.
\newblock In \emph{Transactions of the {{Association}} for {{Computational
  Linguistics}}}, volume~11, pages 1048--1065, Cambridge.

\bibitem[{Cotterell et~al.(2018)Cotterell, Mielke, Eisner, and
  Roark}]{cotterell-etal-2018-are}
Ryan Cotterell, Sebastian~J. Mielke, Jason Eisner, and Brian Roark. 2018.
\newblock \href {https://doi.org/10.18653/v1/N18-2085} {Are all languages
  equally hard to language-model?}
\newblock In \emph{Proceedings of the 2018 {{Conference}} of the {{North
  American Chapter}} of the {{Association}} for {{Computational Linguistics}}:
  {{Human Language}} {{Technologies}}, {{Volume}} 2 ({{Short Papers}})}, pages
  536--541, New Orleans, Louisiana. Association for Computational Linguistics.

\bibitem[{Crispino et~al.(2023)Crispino, Mollica, Astuti, and
  Tardella}]{crispino-etal-2023-efficient}
Marta Crispino, Cristina Mollica, Valerio Astuti, and Luca Tardella. 2023.
\newblock \href {https://doi.org/10.1007/s11222-023-10266-8} {Efficient and
  accurate inference for mixtures of {{Mallows}} models with {{Spearman}}
  distance}.
\newblock \emph{Statistics and Computing}, 33(5):98.

\bibitem[{Croft(2002)}]{croft-2002-typology}
William Croft. 2002.
\newblock \href {https://doi.org/10.1017/CBO9780511840579} {\emph{Typology and
  {{Universals}}}}, 2nd edition.
\newblock Cambridge University Press, Cambridge.

\bibitem[{Di~Marco and Fraser(2024)}]{dimarco-fraser-2024-subword}
Marion Di~Marco and Alexander Fraser. 2024.
\newblock \href {https://doi.org/10.18653/v1/2024.emnlp-main.672} {Subword
  segmentation in {{LLMs}}: {{Looking}} at inflection and consistency}.
\newblock In \emph{Proceedings of the 2024 {{Conference}} on {{Empirical
  Methods}} in {{Natural Language Processing}}}, pages 12050--12060, Miami,
  Florida, USA. Association for Computational Linguistics.

\bibitem[{Diehl~Martinez et~al.(2025)Diehl~Martinez, Africa, Weiss, Salhan,
  Daniels, and Buttery}]{diehlmartinez-etal-2025-pico}
Richard Diehl~Martinez, David~Demitri Africa, Yuval Weiss, Suchir Salhan, Ryan
  Daniels, and Paula Buttery. 2025.
\newblock \href {https://doi.org/10.18653/v1/2025.emnlp-demos.22} {Pico: {{A}}
  modular framework for hypothesis-driven small language model research}.
\newblock In \emph{Proceedings of the 2025 {{Conference}} on {{Empirical
  Methods}} in {{Natural Language Processing}}: {{System Demonstrations}}},
  pages 295--306, Suzhou, China. Association for Computational Linguistics.

\bibitem[{Dryer and Haspelmath(2024)}]{dryer-haspelmath-2024-world}
Matthew Dryer and Martin Haspelmath. 2024.
\newblock \href {https://doi.org/10.5281/ZENODO.13950591} {The {{World Atlas}}
  of {{Language Structures}} online - {{Order}} of subject, object and verb}.

\bibitem[{{El-Naggar} et~al.(2025{\natexlab{a}}){El-Naggar}, Kuribayashi, and
  Briscoe}]{el-naggar-etal-2025-gcgbased}
Nadine {El-Naggar}, Tatsuki Kuribayashi, and Ted Briscoe. 2025{\natexlab{a}}.
\newblock \href {https://doi.org/10.18653/v1/2025.conll-1.35} {{{GCG-based}}
  artificial languages for evaluating inductive biases of neural language
  models}.
\newblock In \emph{Proceedings of the 29th {{Conference}} on {{Computational
  Natural Language Learning}}}, pages 540--556, Vienna, Austria. Association
  for Computational Linguistics.

\bibitem[{{El-Naggar} et~al.(2025{\natexlab{b}}){El-Naggar}, Kuribayashi, and
  Briscoe}]{el-naggar-etal-2025-which}
Nadine {El-Naggar}, Tatsuki Kuribayashi, and Ted Briscoe. 2025{\natexlab{b}}.
\newblock \href {https://doi.org/10.18653/v1/2025.emnlp-main.1803} {Which word
  orders facilitate length generalization in {{LMs}}? {{An}} investigation with
  {{GCG-based}} artificial languages}.
\newblock In \emph{Proceedings of the 2025 {{Conference}} on {{Empirical
  Methods}} in {{Natural Language Processing}}}, pages 35587--35601, Suzhou,
  China. Association for Computational Linguistics.

\bibitem[{Feller(1968)}]{feller-1968-introduction}
William Feller. 1968.
\newblock \emph{An {{Introduction}} to {{Probability Theory}} and its
  {{Applications I}}}, 3rd edition.
\newblock Wiley series in probability and mathematical statistics. Wiley, New
  York.

\bibitem[{Fligner and Verducci(1986)}]{fligner-verducci-1986-distance}
Michael~A. Fligner and Joseph~S. Verducci. 1986.
\newblock \href {https://doi.org/10.1111/j.2517-6161.1986.tb01420.x} {Distance
  based ranking models}.
\newblock \emph{Journal of the Royal Statistical Society: Series B
  (Methodological)}, 48(3):359--369.

\bibitem[{{Franco-S{\'a}nchez} et~al.(2024){Franco-S{\'a}nchez},
  {Mart{\'i}-Llobet}, and {Ferrer-i-Cancho}}]{franco-sanchez-etal-2024-swap}
V{\'i}ctor {Franco-S{\'a}nchez}, Arnau {Mart{\'i}-Llobet}, and Ramon
  {Ferrer-i-Cancho}. 2024.
\newblock \href {https://doi.org/10.48550/arXiv.2404.14192} {Swap distance
  minimization beyond entropy minimization in word order variation}.
\newblock \emph{Preprint}, arXiv:2404.14192.

\bibitem[{Futrell et~al.(2020)Futrell, Gibson, and
  Levy}]{futrell-etal-2020-lossycontext}
Richard Futrell, Edward Gibson, and Roger~P. Levy. 2020.
\newblock \href {https://doi.org/10.1111/cogs.12814} {Lossy-context surprisal:
  {{An}} information-theoretic model of memory effects in sentence processing}.
\newblock \emph{Cognitive Science}, 44(3):e12814.

\bibitem[{Futrell and Mahowald(2025)}]{futrell-mahowald-2025-how}
Richard Futrell and Kyle Mahowald. 2025.
\newblock \href {https://doi.org/10.1017/S0140525X2510112X} {How {{Linguistics
  Learned}} to {{Stop Worrying}} and {{Love}} the {{Language Models}}}.
\newblock \emph{Behavioral and Brain Sciences}, pages 1--98.

\bibitem[{Greenberg(1990)}]{greenberg-1990-universals}
Joseph~H. Greenberg. 1990.
\newblock \href {https://doi.org/10.1515/9781503623217-005} {Some universals of
  grammar with particular reference to the order of meaningful elements}.
\newblock In Keith Denning and Suzanne Kemmer, editors, \emph{On {{Language}}:
  {{Selected Writings}} of {{Joseph H}}. {{Greenberg}}}, pages 40--70. Stanford
  University Press.

\bibitem[{{Gutierrez-Vasques} et~al.(2023){Gutierrez-Vasques}, Bentz, and
  Samard{\v z}i{\'c}}]{gutierrez-vasques-etal-2023-languages}
Ximena {Gutierrez-Vasques}, Christian Bentz, and Tanja Samard{\v z}i{\'c}.
  2023.
\newblock \href {https://doi.org/10.1162/coli_a_00489} {Languages through the
  looking glass of {{BPE}} compression}.
\newblock \emph{Computational Linguistics}, 49(4):943--1001.

\bibitem[{Hahn et~al.(2020)Hahn, Jurafsky, and
  Futrell}]{hahn-etal-2020-universals}
Michael Hahn, Dan Jurafsky, and Richard Futrell. 2020.
\newblock \href {https://doi.org/10.1073/pnas.1910923117} {Universals of word
  order reflect optimization of grammars for efficient communication}.
\newblock \emph{Proceedings of the National Academy of Sciences},
  117(5):2347--2353.

\bibitem[{Hahn and Xu(2022)}]{hahn-xu-2022-crosslinguistic}
Michael Hahn and Yang Xu. 2022.
\newblock \href {https://doi.org/10.1073/pnas.2122604119} {Crosslinguistic word
  order variation reflects evolutionary pressures of dependency and information
  locality}.
\newblock \emph{Proceedings of the National Academy of Sciences},
  119(24):e2122604119.

\bibitem[{Hale(2001)}]{hale-2001-probabilistic}
John Hale. 2001.
\newblock \href {https://doi.org/10.3115/1073336.1073357} {A probabilistic
  {{Earley}} parser as a psycholinguistic model}.
\newblock In \emph{Second meeting of the {{North American Chapter}} of the
  {{Association}} for {{Computational Linguistics}} on {{Language}}
  technologies 2001 - {{NAACL}} '01}, pages 1--8, Pittsburgh, Pennsylvania.
  Association for Computational Linguistics.

\bibitem[{Haley and Wilson(2021)}]{haley-wilson-2021-deep}
Coleman Haley and Colin Wilson. 2021.
\newblock \href {https://aclanthology.org/2021.scil-1.51/} {Deep neural
  networks easily learn unnatural infixation and reduplication patterns}.
\newblock In \emph{Proceedings of the {{Society}} for {{Computation}} in
  {{Linguistics}} 2021}, pages 427--433, Online. Association for Computational
  Linguistics.

\bibitem[{Hammarstr{\"o}m et~al.(2025)Hammarstr{\"o}m, Forkel, Haspelmath, and
  Bank}]{hammarstrom-etal-2025-glottolog}
Harald Hammarstr{\"o}m, Robert Forkel, Martin Haspelmath, and Sebastian Bank.
  2025.
\newblock \href {https://doi.org/10.5281/ZENODO.15525265} {Glottolog 5.2}.

\bibitem[{Harms(1997)}]{harms-1997-estonian}
Robert~T. Harms. 1997.
\newblock \emph{Estonian {{Grammar}}}, 1st edition.
\newblock {Taylor and Francis}, London.

\bibitem[{Harris and Vincent(2012)}]{harris-vincent-2012-romance}
Martin Harris and Nigel Vincent. 2012.
\newblock \emph{Romance {{Languages}}}, 1st edition.
\newblock Routledge {{Language Family Series}}. {Taylor and Francis}, Hoboken.

\bibitem[{Haspelmath(2023)}]{haspelmath-2023-defining}
Martin Haspelmath. 2023.
\newblock \href {https://doi.org/10.1080/00437956.2023.2237272} {Defining the
  word}.
\newblock \emph{WORD}, 69(3):283--297.

\bibitem[{Hawkins(2014)}]{hawkins-2014-crosslinguistic}
John~A. Hawkins. 2014.
\newblock \href {https://doi.org/10.1093/acprof:oso/9780199664993.001.0001}
  {\emph{Cross-{{Linguistic Variation}} and {{Efficiency}}}}, 1st edition.
\newblock Oxford University Press, Oxford.

\bibitem[{Hessel and Schofield(2021)}]{hessel-schofield-2021-how}
Jack Hessel and Alexandra Schofield. 2021.
\newblock \href {https://doi.org/10.18653/v1/2021.acl-short.27} {How effective
  is {{BERT}} without word ordering? {{Implications}} for language
  understanding and data privacy}.
\newblock In \emph{Proceedings of the 59th {{Annual Meeting}} of the
  {{Association}} for {{Computational Linguistics}} and the 11th
  {{International Joint Conference}} on {{Natural Language Processing}}
  ({{Volume}} 2: {{Short Papers}})}, pages 204--211, Online. Association for
  Computational Linguistics.

\bibitem[{Holmes and Hinchliffe(2013)}]{holmes-hinchliffe-2013-swedish}
Philip Holmes and Ian Hinchliffe. 2013.
\newblock \href {https://doi.org/10.4324/9780203381670} {\emph{Swedish: {{A
  Comprehensive Grammar}}}}, 3rd edition.
\newblock Routledge, London.

\bibitem[{Kabatek(2022)}]{kabatek-2022-manual}
Johannes Kabatek. 2022.
\newblock \href {https://doi.org/10.1515/9783110405958} {\emph{Manual of
  {{Brazilian Portuguese Linguistics}}}}, 1st edition.
\newblock Number~21 in Manuals of {{Romance Linguistics}}. Walter de Gruyter
  GmbH, Berlin/Boston.

\bibitem[{Kallini et~al.(2024)Kallini, Papadimitriou, Futrell, Mahowald, and
  Potts}]{kallini-etal-2024-mission}
Julie Kallini, Isabel Papadimitriou, Richard Futrell, Kyle Mahowald, and
  Christopher Potts. 2024.
\newblock \href {https://doi.org/10.18653/v1/2024.acl-long.787} {Mission:
  {{Impossible}} language models}.
\newblock In \emph{Proceedings of the 62nd {{Annual Meeting}} of the
  {{Association}} for {{Computational Linguistics}} ({{Volume}} 1: {{Long
  Papers}})}, pages 14691--14714, Bangkok, Thailand. Association for
  Computational Linguistics.

\bibitem[{Kallini and Potts(2025)}]{kallini-potts-2025-language}
Julie Kallini and Christopher Potts. 2025.
\newblock \href {https://doi.org/10.48550/ARXIV.2512.09394} {Language models as
  tools for investigating the distinction between possible and impossible
  natural languages}.
\newblock \emph{Preprint}, arXiv:2512.09394.

\bibitem[{Karlsson(2017)}]{karlsson-2017-finnish}
Fred Karlsson. 2017.
\newblock \href {https://doi.org/10.4324/9781315743547} {\emph{Finnish: {{A
  Comprehensive Grammar}}}}, 1st edition.
\newblock Routledge, Abingdon, Oxon.

\bibitem[{Katzir(2023)}]{katzir-2023-why}
Roni Katzir. 2023.
\newblock \href {https://doi.org/10.5964/bioling.13153} {Why large language
  models are poor theories of human linguistic cognition: {{A}} reply to
  {{Piantadosi}}}.
\newblock \emph{Biolinguistics}, 17:e13153.

\bibitem[{Kendall(1938)}]{kendall-1938-new}
Maurice~G. Kendall. 1938.
\newblock \href {https://doi.org/10.2307/2332226} {A new measure of rank
  correlation}.
\newblock \emph{Biometrika}, 30(1/2):81.

\bibitem[{Kenesei et~al.(2002)Kenesei, Vago, and
  Fenyvesi}]{kenesei-etal-2002-hungarian}
Istvan Kenesei, Robert~M. Vago, and Anna Fenyvesi. 2002.
\newblock \href {https://doi.org/10.4324/9780203192238} {\emph{Hungarian}}, 1st
  edition.
\newblock Routledge, London.

\bibitem[{Koehn(2005)}]{koehn-2005-europarl}
Philipp Koehn. 2005.
\newblock \href {https://aclanthology.org/2005.mtsummit-papers.11/} {Europarl:
  {{A}} parallel corpus for statistical machine translation}.
\newblock In \emph{Proceedings of {{Machine Translation Summit X}}:
  {{Papers}}}, pages 79--86, Phuket, Thailand.

\bibitem[{Koplenig et~al.(2023)Koplenig, Wolfer, and
  Meyer}]{koplenig-etal-2023-large}
Alexander Koplenig, Sascha Wolfer, and Peter Meyer. 2023.
\newblock \href {https://doi.org/10.1038/s41598-023-42327-3} {A large
  quantitative analysis of written language challenges the idea that all
  languages are equally complex}.
\newblock \emph{Scientific Reports}, 13(1):15351.

\bibitem[{Leivada et~al.(2025)Leivada, Montero, Morosi, Moskvina, Serrano,
  Aguilar, and Guenther}]{leivada-etal-2025-large}
Evelina Leivada, Raquel Montero, Paolo Morosi, Natalia Moskvina, Tamara
  Serrano, Marcel Aguilar, and Fritz Guenther. 2025.
\newblock \href {https://doi.org/10.48550/ARXIV.2509.15114} {Large language
  model probabilities cannot distinguish between possible and impossible
  language}.
\newblock \emph{Preprint}, arXiv:2509.15114.

\bibitem[{Levshina et~al.(2023)Levshina, Namboodiripad,
  {Allassonni{\`e}re-Tang}, Kramer, Talamo, Verkerk, Wilmoth, Rodriguez,
  Gupton, Kidd, Liu, Naccarato, Nordlinger, Panova, and
  Stoynova}]{levshina-etal-2023-why}
Natalia Levshina, Savithry Namboodiripad, Marc {Allassonni{\`e}re-Tang}, Mathew
  Kramer, Luigi Talamo, Annemarie Verkerk, Sasha Wilmoth, Gabriela~Garrido
  Rodriguez, Timothy~Michael Gupton, Evan Kidd, Zoey Liu, Chiara Naccarato,
  Rachel Nordlinger, Anastasia Panova, and Natalia Stoynova. 2023.
\newblock \href {https://doi.org/10.1515/ling-2021-0098} {Why we need a
  gradient approach to word order}.
\newblock \emph{Linguistics}, 61(4):825--883.

\bibitem[{Levy(2008)}]{levy-2008-expectationbased}
Roger Levy. 2008.
\newblock \href {https://doi.org/10.1016/j.cognition.2007.05.006}
  {Expectation-based syntactic comprehension}.
\newblock \emph{Cognition}, 106(3):1126--1177.

\bibitem[{Liu et~al.(2025)Liu, Yan, and Liu}]{liu-etal-2025-complexity}
Siqi Liu, Jianwei Yan, and Haitao Liu. 2025.
\newblock \href {https://doi.org/10.1515/zrp-2025-0032} {The complexity
  trade-off between morphological richness and word order freedom in
  {{Romance}} languages: {{A}} quantitative perspective}.
\newblock \emph{Zeitschrift f\"ur romanische Philologie}, 141(2):323--349.

\bibitem[{{Lundskaer-Nielsen} and
  Holmes(2015)}]{lundskaer-nielsen-holmes-2015-danish}
Tom {Lundskaer-Nielsen} and Philip Holmes. 2015.
\newblock \href {https://doi.org/10.4324/9780203853023} {\emph{Danish: {{A
  Comprehensive Grammar}}}}, 2nd edition.
\newblock Routledge, London.

\bibitem[{MacKay(2019)}]{mackay-2019-information}
David J.~C. MacKay. 2019.
\newblock \emph{Information {{Theory}}, {{Inference}}, and {{Learning
  Algorithms}}}, 22nd edition.
\newblock Cambridge University Press, Cambridge.

\bibitem[{Mallows(1957)}]{mallows-1957-nonnull}
Colin~L. Mallows. 1957.
\newblock \href {https://doi.org/10.2307/2333244} {Non-null ranking models.
  {{I}}}.
\newblock \emph{Biometrika}, 44(1/2):114--130.

\bibitem[{Mielke et~al.(2019)Mielke, Cotterell, Gorman, Roark, and
  Eisner}]{mielke-etal-2019-what}
Sebastian~J. Mielke, Ryan Cotterell, Kyle Gorman, Brian Roark, and Jason
  Eisner. 2019.
\newblock \href {https://doi.org/10.18653/v1/P19-1491} {What kind of language
  is hard to language-model?}
\newblock In \emph{Proceedings of the 57th {{Annual Meeting}} of the
  {{Association}} for {{Computational Linguistics}}}, pages 4975--4989,
  Florence, Italy. Association for Computational Linguistics.

\bibitem[{Naughton(2008)}]{naughton-2008-czech}
James~D. Naughton. 2008.
\newblock \emph{{Czech: An Essential Grammar}}, 2nd edition.
\newblock {Essential grammars}. Routledge, London.

\bibitem[{Nijs et~al.(2025)Nijs, {Van de Velde}, and
  Cuyckens}]{nijs-etal-2025-word}
Julie Nijs, Freek {Van de Velde}, and Hubert Cuyckens. 2025.
\newblock \href {https://doi.org/10.3390/e27010053} {Is word order responsive
  to morphology? {{Disentangling}} cause and effect in morphosyntactic change
  in five {{Western European}} languages}.
\newblock \emph{Entropy}, 27(1):53.

\bibitem[{Papadimitriou et~al.(2022)Papadimitriou, Futrell, and
  Mahowald}]{papadimitriou-etal-2022-when}
Isabel Papadimitriou, Richard Futrell, and Kyle Mahowald. 2022.
\newblock \href {https://doi.org/10.18653/v1/2022.acl-short.71} {When
  classifying grammatical role, {{BERT}} doesn't care about word order...
  except when it matters}.
\newblock In \emph{Proceedings of the 60th {{Annual Meeting}} of the
  {{Association}} for {{Computational Linguistics}} ({{Volume}} 2: {{Short
  Papers}})}, pages 636--643, Dublin, Ireland. Association for Computational
  Linguistics.

\bibitem[{Papadimitriou and
  Prince(2025)}]{papadimitriou-prince-2025-vocabulary}
Isabel Papadimitriou and Jacob Prince. 2025.
\newblock \href {https://doi.org/10.48550/arXiv.2510.07613} {Vocabulary
  embeddings organize linguistic structure early in language model training}.
\newblock \emph{Preprint}, arXiv:2510.07613.

\bibitem[{Pham et~al.(2021)Pham, Bui, Mai, and Nguyen}]{pham-etal-2021-out}
Thang Pham, Trung Bui, Long Mai, and Anh Nguyen. 2021.
\newblock \href {https://doi.org/10.18653/v1/2021.findings-acl.98} {Out of
  order: {{How}} important is the sequential order of words in a sentence in
  {{Natural Language Understanding}} tasks?}
\newblock In \emph{Findings of the {{Association}} for {{Computational
  Linguistics}}: {{ACL-IJCNLP}} 2021}, pages 1145--1160, Online. Association
  for Computational Linguistics.

\bibitem[{Piantadosi(2014)}]{piantadosi-2014-zipfs}
Steven~T. Piantadosi. 2014.
\newblock \href {https://doi.org/10.3758/s13423-014-0585-6} {Zipf's word
  frequency law in natural language: {{A}} critical review and future
  directions}.
\newblock \emph{Psychonomic Bulletin \& Review}, 21(5):1112--1130.

\bibitem[{Piantadosi(2024)}]{piantadosi-2024-modern}
Steven~T. Piantadosi. 2024.
\newblock \href {https://doi.org/10.5281/ZENODO.12665933} {Modern language
  models refute {{Chomsky}}'s approach to language}.
\newblock In \emph{From {{Fieldwork}} to {{Linguistic Theory}}}, 1st edition,
  number~15 in Empirically {{Oriented Theoretical Morphology}} and {{Syntax}},
  pages 353--414. Language Science Press, Berlin.

\bibitem[{Piantadosi et~al.(2012)Piantadosi, Tily, and
  Gibson}]{piantadosi-etal-2012-communicative}
Steven~T. Piantadosi, Harry Tily, and Edward Gibson. 2012.
\newblock \href {https://doi.org/10.1016/j.cognition.2011.10.004} {The
  communicative function of ambiguity in language}.
\newblock \emph{Cognition}, 122(3):280--291.

\bibitem[{Ploeger et~al.(2024)Ploeger, Poelman, {de Lhoneux}, and
  Bjerva}]{ploeger-etal-2024-what}
Esther Ploeger, Wessel Poelman, Miryam {de Lhoneux}, and Johannes Bjerva. 2024.
\newblock \href {https://doi.org/10.18653/v1/2024.emnlp-main.326} {What is
  ``typological diversity'' in {{NLP}}?}
\newblock In \emph{Proceedings of the 2024 {{Conference}} on {{Empirical
  Methods}} in {{Natural Language Processing}}}, pages 5681--5700, Miami,
  Florida, USA. Association for Computational Linguistics.

\bibitem[{Ploeger et~al.(2025)Ploeger, Poelman, {H{\o}eg-Petersen},
  Schlichtkrull, De~Lhoneux, and Bjerva}]{ploeger-etal-2025-principled}
Esther Ploeger, Wessel Poelman, Andreas~Holck {H{\o}eg-Petersen}, Anders
  Schlichtkrull, Miryam De~Lhoneux, and Johannes Bjerva. 2025.
\newblock \href {https://doi.org/10.1162/COLI.a.577} {A principled framework
  for evaluating on typologically diverse languages}.
\newblock \emph{Computational Linguistics}, pages 1--36.

\bibitem[{Poelman et~al.(2025)Poelman, Bauwens, and {de
  Lhoneux}}]{poelman-etal-2025-confounding}
Wessel Poelman, Thomas Bauwens, and Miryam {de Lhoneux}. 2025.
\newblock \href {https://doi.org/10.18653/v1/2025.emnlp-main.369} {Confounding
  factors in relating model performance to morphology}.
\newblock In \emph{Proceedings of the 2025 {{Conference}} on {{Empirical
  Methods}} in {{Natural Language Processing}}}, pages 7273--7298, Suzhou,
  China. Association for Computational Linguistics.

\bibitem[{Ponti et~al.(2019)Ponti, O'Horan, Berzak, Vuli{\'c}, Reichart,
  Poibeau, Shutova, and Korhonen}]{ponti-etal-2019-modeling}
Edoardo~Maria Ponti, Helen O'Horan, Yevgeni Berzak, Ivan Vuli{\'c}, Roi
  Reichart, Thierry Poibeau, Ekaterina Shutova, and Anna Korhonen. 2019.
\newblock \href {https://doi.org/10.1162/coli_a_00357} {Modeling language
  variation and universals: {{A}} survey on typological linguistics for natural
  language processing}.
\newblock \emph{Computational Linguistics}, 45(3):559--601.

\bibitem[{Praulin{\v s}(2012)}]{praulins-2012-latvian}
Dace Praulin{\v s}. 2012.
\newblock \href {https://doi.org/10.4324/9780203124420} {\emph{Latvian: {{An
  Essential Grammar}}}}, 1st edition.
\newblock Essential grammars. Routledge, London.

\bibitem[{Sampson(2009)}]{sampson-2009-linguistic}
Geoffrey Sampson. 2009.
\newblock \href {https://doi.org/10.1093/oso/9780199545216.003.0001} {A
  linguistic axiom challenged}.
\newblock In Geoffrey Sampson, David Gil, and Peter Trudgill, editors,
  \emph{Language {{Complexity}} as an {{Evolving Variable}}}, pages 1--18.
  Oxford University PressOxford.

\bibitem[{Sampson et~al.(2009)Sampson, Gil, and
  Trudgill}]{sampson-etal-2009-language}
Geoffrey Sampson, David Gil, and Peter Trudgill, editors. 2009.
\newblock \href {https://doi.org/10.1093/oso/9780199545216.001.0001}
  {\emph{Language {{Complexity}} as an {{Evolving Variable}}}}.
\newblock Oxford University PressOxford, New York.

\bibitem[{Schad et~al.(2010)Schad, Nuthmann, and Engbert}]{schad-etal-2010-eye}
Daniel~J. Schad, Antje Nuthmann, and Ralf Engbert. 2010.
\newblock \href {https://doi.org/10.1016/j.visres.2010.08.005} {Eye movements
  during reading of randomly shuffled text}.
\newblock \emph{Vision Research}, 50(23):2600--2616.

\bibitem[{Sch{\"u}rmann and
  Grassberger(1996)}]{schurmann-grassberger-1996-entropy}
Thomas Sch{\"u}rmann and Peter Grassberger. 1996.
\newblock \href {https://doi.org/10.1063/1.166191} {Entropy estimation of
  symbol sequences}.
\newblock \emph{Chaos: An Interdisciplinary Journal of Nonlinear Science},
  6(3):414--427.

\bibitem[{Shani et~al.(2026)Shani, Reif, Roll, Jurafsky, and
  Shutova}]{shani-etal-2026-roots}
Chen Shani, Yuval Reif, Nathan Roll, Dan Jurafsky, and Ekaterina Shutova. 2026.
\newblock \href {https://doi.org/10.48550/arXiv.2601.07220} {The roots of
  performance disparity in multilingual language models: {{Intrinsic}} modeling
  difficulty or design choices?}
\newblock \emph{Preprint}, arXiv:2601.07220.

\bibitem[{Shannon(1948)}]{shannon-1948-mathematical}
Claude~E. Shannon. 1948.
\newblock \href {https://doi.org/10.1002/j.1538-7305.1948.tb01338.x} {A
  mathematical theory of communication}.
\newblock \emph{Bell System Technical Journal}, 27(3):379--423.

\bibitem[{Siewierska(2010)}]{siewierska-2010-constituent}
Anna Siewierska, editor. 2010.
\newblock \href {https://doi.org/10.1515/9783110812206} {\emph{Constituent
  {{Order}} in the {{Languages}} of {{Europe}}}}, 1st edition.
\newblock Number Eurotyp 20-1 in Empirical {{Approaches}} to {{Language
  Typology}}. De Gruyter, Berlin.

\bibitem[{Smith and Levy(2013)}]{smith-levy-2013-effect}
Nathaniel~J. Smith and Roger Levy. 2013.
\newblock \href {https://doi.org/10.1016/j.cognition.2013.02.013} {The effect
  of word predictability on reading time is logarithmic}.
\newblock \emph{Cognition}, 128(3):302--319.

\bibitem[{Someya et~al.(2025)Someya, Svete, DuSell, O'Donnell, Giulianelli, and
  Cotterell}]{someya-etal-2025-information}
Taiga Someya, Anej Svete, Brian DuSell, Timothy~J. O'Donnell, Mario
  Giulianelli, and Ryan Cotterell. 2025.
\newblock \href {https://doi.org/10.18653/v1/2025.acl-long.1357} {Information
  locality as an inductive bias for neural language models}.
\newblock In \emph{Proceedings of the 63rd {{Annual Meeting}} of the
  {{Association}} for {{Computational Linguistics}} ({{Volume}} 1: {{Long
  Papers}})}, pages 27995--28013, Vienna, Austria. Association for
  Computational Linguistics.

\bibitem[{Sulkala and Karjalainen(2012)}]{sulkala-karjalainen-2012-finnish}
Helena Sulkala and Merja Karjalainen. 2012.
\newblock \href
  {https://www.routledge.com/Finnish/Karalainen-Sulkala/p/book/9780415657136}
  {\emph{Finnish}}, 1st edition.
\newblock Descriptive grammars. Routledge, London.

\bibitem[{Svenonius(2025)}]{svenonius-2025-word}
Peter Svenonius. 2025.
\newblock \href {https://doi.org/10.1146/annurev-linguistics-011724-121650}
  {Word order universals and their relationship to structure}.
\newblock \emph{Annual Review of Linguistics}, 11(1):137--162.

\bibitem[{Tang(2019)}]{tang-2019-mallows}
Wenpin Tang. 2019.
\newblock \href {https://proceedings.mlr.press/v97/tang19a.html} {Mallows
  ranking models: {{Maximum}} likelihood estimate and regeneration}.
\newblock In \emph{Proceedings of the 36th {{International Conference}} on
  {{Machine Learning}}}, volume~97 of \emph{Proceedings of {{Machine Learning
  Research}}}, pages 6125--6134. Proceedings of Machine Learning Research.

\bibitem[{Tao et~al.(2024)Tao, Liu, Dou, Muennighoff, Wan, Luo, Lin, and
  Wong}]{tao-etal-2024-scaling}
Chaofan Tao, Qian Liu, Longxu Dou, Niklas Muennighoff, Zhongwei Wan, Ping Luo,
  Min Lin, and Ngai Wong. 2024.
\newblock \href
  {https://proceedings.neurips.cc/paper_files/paper/2024/hash/cf5a019ae9c11b4be88213ce3f85d85c-Abstract-Conference.html}
  {Scaling laws with vocabulary: {{Larger}} models deserve larger
  vocabularies}.
\newblock In \emph{Advances in {{Neural Information Processing System}}},
  volume~37, pages 114147--114179. Curran Associates, Inc.

\bibitem[{Xu et~al.(2025)Xu, Kuribayashi, Oseki, Cotterell, and
  Warstadt}]{xu-etal-2025-can}
Tianyang Xu, Tatsuki Kuribayashi, Yohei Oseki, Ryan Cotterell, and Alex
  Warstadt. 2025.
\newblock \href {https://doi.org/10.48550/arXiv.2502.12317} {Can language
  models learn typologically implausible languages?}
\newblock \emph{Preprint}, arXiv:2502.12317.

\bibitem[{Yang et~al.(2025)Yang, Aoyama, Yao, and
  Wilcox}]{yang-etal-2025-anything}
Xiulin Yang, Tatsuya Aoyama, Yuekun Yao, and Ethan Wilcox. 2025.
\newblock \href {https://doi.org/10.18653/v1/2025.acl-long.1264} {Anything
  goes? {{A}} crosslinguistic study of (im)possible language learning in
  {{LMs}}}.
\newblock In \emph{Proceedings of the 63rd {{Annual Meeting}} of the
  {{Association}} for {{Computational Linguistics}} ({{Volume}} 1: {{Long
  Papers}})}, pages 26058--26077, Vienna, Austria. Association for
  Computational Linguistics.

\bibitem[{Zipf(1935)}]{zipf-1935-psychobiology}
George~Kingsley Zipf. 1935.
\newblock \emph{The {{Psycho-Biology}} of {{Language}}: {{An Introduction}} to
  {{Dynamic Philology}}}, 1st edition.
\newblock Houghton Mifflin, Boston, Massachusetts.

\bibitem[{Zipf(1949)}]{zipf-1949-human}
George~Kingsley Zipf. 1949.
\newblock \href
  {https://www.mpi.nl/publications/item2407822/human-behavior-and-principle-least-effort-introduction-human-eoclogy}
  {\emph{Human {{Behavior}} and the {{Principle}} of {{Least Effort}}: {{An
  Introduction}} to {{Human Ecology}}}}.
\newblock Addison-Wesley Press., Cambridge.

\bibitem[{Ziv et~al.(2025)Ziv, Lan, Chemla, and
  Katzir}]{ziv-etal-2025-biasless}
Imry Ziv, Nur Lan, Emmanuel Chemla, and Roni Katzir. 2025.
\newblock \href {https://doi.org/10.48550/ARXIV.2510.07178} {Biasless language
  models learn unnaturally: {{How LLMs}} fail to distinguish the possible from
  the impossible}.
\newblock \emph{Preprint}, arXiv:2510.07178.

\end{thebibliography}
\end{document}